\pdfoutput=1
\documentclass[11pt]{article}
\usepackage[]{emnlp2021}
\usepackage{hyperref}
\usepackage{times}
\usepackage[T1]{fontenc}
\usepackage[utf8]{inputenc}
\usepackage{microtype}
\usepackage{comment}
\usepackage{multicol}
\usepackage{latexsym}

\usepackage{enumitem,microtype}
\usepackage{mdframed}
\usepackage{amsmath}
\usepackage{amssymb}
\usepackage{todonotes}
\usepackage{graphicx}
\usepackage{caption,subcaption}
\usepackage{booktabs}
\usepackage{graphicx}
\usepackage{relsize}
\usepackage{latexsym}
\usepackage{paralist}
\usepackage{times}
\usepackage{url}
\usepackage{multirow}
\usepackage[normalem]{ulem}
\usepackage{microtype}

\newenvironment{itemize*}%
  {\begin{itemize}%
    \setlength{\itemsep}{0.9pt}%
    \setlength{\parskip}{0.9pt}%
    \setlength{\topsep}{0.9pt}}%
  {\end{itemize}}

\newcommand*{\affaddr}[1]{#1} %
\newcommand*{\affmark}[1][*]{\textsuperscript{#1}}
\newcommand*{\email}[1]{\texttt{#1}}

\title{Generalization in NLI: Ways (Not) To Go Beyond Simple Heuristics}

\author{
	Prajjwal Bhargava\affmark[1], Aleksandr Drozd\affmark[2,3], Anna Rogers\affmark[3,4]\\	
	\affaddr{\affmark[1] \normalsize The University of Texas at Dallas}\\
	\affaddr{\affmark[2] \normalsize RIKEN Center for Computational Science} \\
	\affaddr{\affmark[3] \normalsize Tokyo Institute of Technology} \\
	\affaddr{\affmark[4] \normalsize Center for Social Data Science, University of Copenhagen} \\
	\affmark[1]{\normalsize \email{prajjwalin@protonmail.com}}, \hspace{0.cm} 
	\affmark[2]{\normalsize \email{alex@blackbird.pw}}, \hspace{0.cm}
	\affmark[3]{\normalsize \email{arogers@sodas.ku.dk}}
}

\usepackage[nameinlink]{cleveref}

\crefformat{section}{\S#2#1#3} %
\crefformat{subsection}{\S#2#1#3}
\crefformat{subsubsection}{\S#2#1#3}
\crefformat{paragraph}{\S#2#1#3}

\begin{document}
\maketitle

\begin{abstract}

Much of recent progress in NLU was shown to be due to models' learning dataset-specific heuristics. We conduct a case study of generalization in NLI (from MNLI to the adversarially constructed HANS dataset) in a range of BERT-based architectures (adapters, Siamese Transformers, HEX debiasing), as well as with subsampling the data and increasing the model size. We report 2 successful and 3 unsuccessful strategies, all providing insights into how Transformer-based models learn to generalize.

\end{abstract}

\section{Introduction}
\label{sec:intro}

Many popular NLP datasets contain spurious patterns, which get learned instead of the actual task \cite{gururangan-etal-2018-annotation,belinkov-etal-2019-dont,RogersKovalevaEtAl_2020_Getting_Closer_to_AI_Complete_Question_Answering_Set_of_Prerequisite_Real_Tasks,GardnerMerrillEtAl_2021_Competency_Problems_On_Finding_and_Removing_Artifacts_in_Language_Data}. This raises the issue of %
learning methods that would avoid that problem. We present a case study of generalization to adversarial data in Natural Language Inference (NLI), reporting both positive and negative results for a range of BERT-based approaches.%

\section{Methodology}

\begin{figure}[!t]
\centering
    \includegraphics[height=0.65\linewidth, scale=0.14]{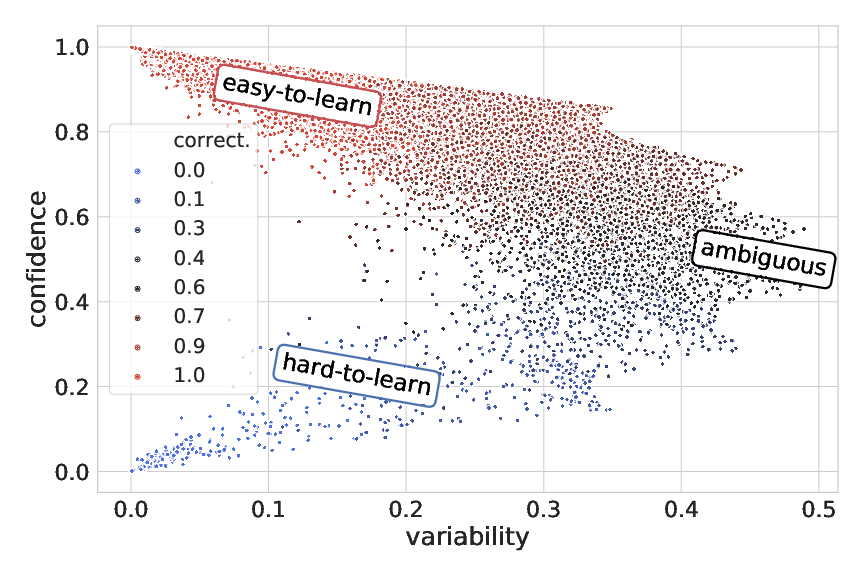}
  \caption{MNLI data map (with RoBERTa-large)}
\label{fig:data_map_roberta}
\end{figure}

\textbf{Data.} NLI is a 3-class classification task: does the premise entails, contradicts, or is neutral with respect to the hypothesis? MNLI \cite{N18-1101} is one of the most popular resources for this task, but it has been shown to suffer from both annotation artifacts \cite{gururangan-etal-2018-annotation, poliak-etal-2018-hypothesis} and annotator bias \cite{GevaGoldbergEtAl_2019_Are_We_Modeling_Task_or_Annotator_Investigation_of_Annotator_Bias_in_Natural_Language_Understanding_Datasetsa}. A cartography \cite{swayamdipta-etal-2020-dataset} map of MNLI (\cref{fig:data_map_roberta}) suggests that most of its examples are easy to learn, which explains why vanilla fine-tuning with modern models is sufficient to achieve high accuracy on MNLI benchmark.

We measure generalization of models fine-tuned on MNLI with HANS \cite{mccoy-etal-2019-right}, a synthetic dataset targeting  \textit{lexical overlap, subsequence} and \textit{constituent} heuristics. According to \citet{mccoy-etal-2019-right}, a model trained on MNLI is likely to learn these heuristics and thus predict the ``entailment" label for most HANS examples. E.g. it would incorrectly predict that ``\textit{The doctor was paid by the actor}" entails ``\textit{The doctor paid the actor}", simply because these sentences contain the same words. See \autoref{app:data} for more examples. %

\textbf{Methodology.}\label{experimental_setup}
We experiment with variants\footnote{Most models we used were provided by HuggingFace Transformers library. In scope of this project we ported the smaller BERT versions by \citet{turc2019wellread} for that library.} of BERT \cite{devlin-etal-2019-bert}, RoBERTa \cite{DBLP:journals/corr/abs-1907-11692}, and ALBERT \cite{Lan2020ALBERT:}. Our implementation\footnote{\url{https://github.com/prajjwal1/generalize_lm_nli}} is based on Transformers \cite{Wolf2019HuggingFacesTS} and Pytorch \cite{NEURIPS2019_9015}, and for two experiments we also report results with a custom Pytorch-Lightning trainer\footnote{\url{https://github.com/vecto-ai/langmo}}. HANS has 30K examples used only for testing, where we report the accuracy\footnote{Since HANS contains only two labels (entailment, non-entailment), and a model trained on MNLI would have three (entailment, contradiction, neutral), a completely random model would be biased towards the ``non-entailment". For direct comparison with MNLI we report the average accuracy for the two HANS labels, unless specified otherwise.
}. 
MNLI test set is not public, and we report accuracy on the ``matched'' dev set (20K examples, 393K for training). %

\section{Experiments}
\label{sec:sampling}

\begin{table*}[!h]
\footnotesize
	\centering
	\begin{tabular}{p{2.4cm}p{3.7cm}|
	    p{1.3cm}p{1.3cm}p{.6cm}|
	    p{1.3cm}p{1.3cm}p{.6cm}}
		\toprule
		\multirow{2}{*}{Architecture} & \multirow{2}{*}{Encoder} & \multicolumn{3}{c}{HF trainer}  & \multicolumn{3}{c}{Custom Trainer} \\	
		 &  & MNLI/std & HANS/std & runs & MNLI/std & HANS/std & runs\\ \midrule
		\multirow{2}{*}{\shortstack[1]{Siamese networks /\\ frozen encoder}} 
        & BERT-base  & 51.43 & 50.74 & 1 & 57.2 / 0.2 & 51.3 / 0.1 & 3\\ %
        & BERT-large & 51.72 & 51.12 & 1 & 61.4 / 0.1& 51.6 / 0.1& 5 \\ \hline 
		\multirow{3}{*}{\shortstack[1]{Siamese networks /\\ trainable encoder}} 
		 & BERT-base & 58.9 & 52.79 & 1 & 76.5 / 0.03 & 51.3 / 0.03 & 3\\ %
		 & BERT-large   & 59.9 & 51.21 & 1 & 78.7 & 52.5 & 1 \\ 
		\hline 
		\multirow{6}{*}{Adapter networks} 
		 & BERT-base & 82.6 & 50.97 & 1\\
		 & BERT-large & 84.75 & 57.17 & 1\\
		 & RoBERTa-base & 86.33 & 57.21 & 1\\
		 & RoBERTa-large & 90.4 & 75.93 & 1\\
		\hline
		HEX debiasing & BERT-base & 56.25 & 50.58 & 1\\
		\hline

\multirow{10}{2.4cm}{Vanilla finetuning: increased model size} 
    & BERT-tiny (4.4M)     & 64.48/0.24 & 50/0 & 3 & 67.4 / 0.2 & 50 / 0.02 & 5\\
	& BERT-mini (11.3M)    & 72.3/0.29 & 50.97/0.04 & 3 & 76.3 / 1& 52.3 / 0.3 & 10\\
	& BERT-small (29.1M)   & 76.48/0.12 & 50.39/0.14 & 3 & 78.4 / 0.5 & 51.1 / 0.3 & 5\\
	& BERT-medium (41.7M)  & 79.64/0.14 & 51.02/0.26 & 3 & 80 / 0.3 & 52 / 0.4& 5\\
	& BERT-base (110M)     & 83.74/0.04 & 53.98/0.78 & 3 & 84 / 0.2 & 69 / 4 & 16\\
	& BERT-large (340M)    & 85.9/0.02 & 72.04/1.97 & 3 & 86.5 / 0.1 & 77.8 / 2.4 & 3 \\
	& RoBERTa-base (125M)  & 87.46/0.1 & 73.11/1.13 & 3 & 87.5 / 0.3 & 77.7 / 1.7 &10\\
	& RoBERTa-large (355M) & 90.3/0.07 & 79.95/0.56 & 3 & 90   / 0.4 & 82.05 / 1 & 3\\
	& ALBERT-base-v2 (11M) & 83.06/0.13 & 66.6/0.78 & 3 & 84.2 / 0.6 & 69.2 / 2.2 & 4 \\
	& ALBERT-large-v2 (17M)& 85.08/0.3 & 70.64/2.91 & 3 & 85.5 / 0.9 & 70.5 / 1.6 & 4 \\
		\bottomrule
	\end{tabular}
	\caption{Generalization from MNLI to HANS in selected approaches.}
	\label{tab:results-algorithmic}
\end{table*}

There are two main directions for solving the generalization challenge: modifying the training distribution and the model itself. For the former we experimented with subsampling (\ref{sec: cartography}), and for the latter -- with 
bottlenecking with Siamese Transformers (\ref{siamese_transformers}) and adapters (\ref{adapters}), explicit debiasing (\ref{hex_}), and increasing model size (\ref{larger_models}). This section presents the motivation and setup for all experiments, and all the results %
are shown in \cref{sec:results}.

\paragraph{Subsampling the training data with cartography.} \label{sec: cartography}
Data cartography \cite{swayamdipta-etal-2020-dataset} characterizes training data points via the model’s confidence in the true class, and the variability of
this confidence across epochs. \Cref{fig:data_map_roberta} shows that MNLI examples form a spectrum: some are consistently ``easy'' (high-confidence) and ``hard'' (low-confidence) across epochs. ``Ambiguous'' samples have midrange confidence and high variability. If much of MNLI is ``easy'', presumably these samples are less informative.

\textit{Experiments.} We partition MNLI based on the training dynamics of RoBERTa-large and BERT-base, and train the respective models on varying amounts of ``hard'' and ``ambiguous'' examples (preceded by 25\% of ``easy'' samples for 2 epochs). See \cref{sec:cartography-explained} for more details. %

\paragraph{Siamese Networks.}
\label{siamese_transformers}
In this architecture predictions are based on a pair of inputs \cite{10.1109/CVPR.2005.202, koch2015siamese}. It was successful on NLI \cite{chen-etal-2017-enhanced} and in conjunction with BERT encoders \cite{reimers-gurevych-2019-sentence}. One of their properties is forcing the model to consider the relation between two sequences in a more holistic way than with cross-attention between concatenated premise+hypothesis (as in standard BERT fine-tuning). %
Intuitively, encoding premise and hypothesis separately could bottleneck\footnote{
The information bottleneck idea \cite{TishbyPereiraEtAl_2000_information_bottleneck_method,AlemiFischerEtAl_2016_Deep_Variational_Information_Bottleneck} has recently been successfully adapted for BERT fine-tuning to avoid overfitting in a low-resource setting by \citet{MahabadiBelinkovEtAl_2020_Variational_Information_Bottleneck_for_Effective_Low-Resource_Fine-Tuning}, who propose a regularization term suppressing the learning of irrelevant information.
} their interaction and encourage learning more abstract patterns, which is what we need here: ideally, an NLI model would learn logical rules rather than numerous lexical or syntactic patterns.

\textit{Experiments.} Our Siamese Transformer consists of a MLP and two BERT encoders which receive hypotheses and premises in a segregated manner. 
We used mean-pooled outputs of last transformer layer (CLS embedding yielded similar results) combined as $[U, V, U-V, U*V]$ as inputs to MLP classifier. 
We experiment with base and large BERTs, with both frozen and trainable encoders. %

\paragraph{Adapter Networks.}
\label{adapters}
Intuitively, standard fine-tuning of BERT changes the amount of task-independent linguistic knowledge that the model can store, and may corrupt it (if the supervised task has significant artifacts). Therefore, by adding separate task-specific components rather than changing the language model weights, we could expect to increase the amount of non-task-specific knowledge in the model. This could be done with adapters \cite{pmlr-v97-houlsby19a,pfeiffer2020adapterfusion}: trainable MLPs inserted within each sub-layer of encoder. Concretely, in a transformer layer $l$, additional adapter parameters $\phi_{l}$ are introduced to learn task specific parameters while keeping pre-trained parameters intact. Having smaller trainable components should also bottleneck the model and encourage it to learn higher-level patterns.

\textit{Experiments.} 
We add adapters in each sub-layer as proposed in \citet{pmlr-v97-houlsby19a} to BERT and RoBERTa with the configuration defined in \citet{pfeiffer2020adapterfusion}. The adapter consists of two linear layers (up and down) with a bottleneck of reduction factor of 16 and the ReLU non-linearity.

\paragraph{Explicit De-biasing.}%
\label{hex_}
If MNLI `teaches' to rely on superficial features, we could try to avoid them. Following \citet{zhou-bansal-2020-towards}, we use HEX projection \cite{wang2018learning}. %
The system includes the main Transformer encoder and a `naive' model learning superficial features. HEX orthogonally projects the Transformer representation into the affine space the most different from the `naive' representation, hopefully removing the bias.

\textit{Experiments.} We extract pooled representations from our main model (BERT-base). The `naive' model is a CBOW model with a self-attention layer \cite{NIPS2017_7181} to capture co-occurrence information from the sequence with input and token embeddings. See \citet{wang2018learning} for more details on the method, and \autoref{appendix:hex_appendix} for implementation and hyperparameter details. During inference, we use logits from BERT only.

\paragraph{Increasing Model Size.}\label{larger_models}
Scaling language models to massive amounts of data has been a reliable source of success on NLP leaderboards, and yielded some interesting emergent properties \cite{brown2020language, JMLR:v21:20-074}. If pre-training ``teaches" transferable linguistic knowledge, the models absorbing more data could be expected to generalize better.   

\textit{Experiments.} We perform standard fine-tuning on MNLI with variants of BERT: tiny, mini, small, medium by \citet{turc2019wellread}, base and large by \citet{devlin-etal-2019-bert}, as well as RoBERTa \cite{DBLP:journals/corr/abs-1907-11692} and ALBERT \cite{Lan2020ALBERT:}. In this and the Siamese network experiment we report not only the results obtained with the HuggingFace Trainer, but also with our custom implementation based on Pytorch Lightning (also with the AdamW optimizer and with similar learning rates). %

\section{Results and Discussion}
\label{sec:results}

\subsection{Negative Results}

\autoref{tab:results-algorithmic} shows that \textbf{Siamese networks and HEX debiasing perform at chance level on HANS. Adapters work better, but do not match vanilla fine-tuning of their base models}. While it is impossible to prove the negative, our experience suggests that, given a reasonable amount of effort, these approaches are not the most promising for the generalization problem we considered. The paper is accompanied by code for our implementations.

Our Siamese model would be expected to fail if high performance of vanilla BERT was largely due to  cross-attention across [premise + hypothesis], enabling it to learn many specific patterns (such as negation in the hypothesis). %
Our bottleneck MLP would not have the capacity to do that, and it clearly also fails to find a more abstract pattern in the representations it receives. Further experiments are needed to verify this hypothesis. Whether or not overall we would like our NLI models to be able to operate with independent representations of premise and hypothesis rather than cross-attention within one representation, is an open question.%

For HEX, \citet{zhou-bansal-2020-towards} suggest that the problem might be that it has access only to the final output of BERT, which could contain more information about the predicted NLI labels than the input sequence as such. Then there would be little to debias. Our results support this hypothesis, but more qualitative research is needed to verify it.%

The RoBERTa-large MNLI results of our adapter implementation is on par with the recent state-of-the-art Compacter adapters on T5 \cite{mahabadi2021compacter}, but generalization in both BERT and RoBERTa is overall worse than with vanilla fine-tuning. Following on the recent report of adapter efficacy in low-resource setting \cite{HeLiuEtAl_2021_On_Effectiveness_of_Adapter-based_Tuning_for_Pretrained_Language_Model_Adaptation}, we conducted an additional experiment with adapters and RoBERTa-large, where the model had to learn from a small, more informative subsample.  %
At 1024 training examples adapters performed better when the MNLI subsample was diverse (selected with K-means-based clustering, see \cref{sec:clustering}) rather than randomly selected:  80.7\% vs 85\%. But the generalization to HANS was still not very impressive: 67.8\% vs 57.5\%, respectively. This strategy does seem to select more informative examples for MNLI distribution, but not for HANS.

\begin{figure}[!t]
\centering
    \includegraphics[height=0.5\linewidth, scale=0.1]{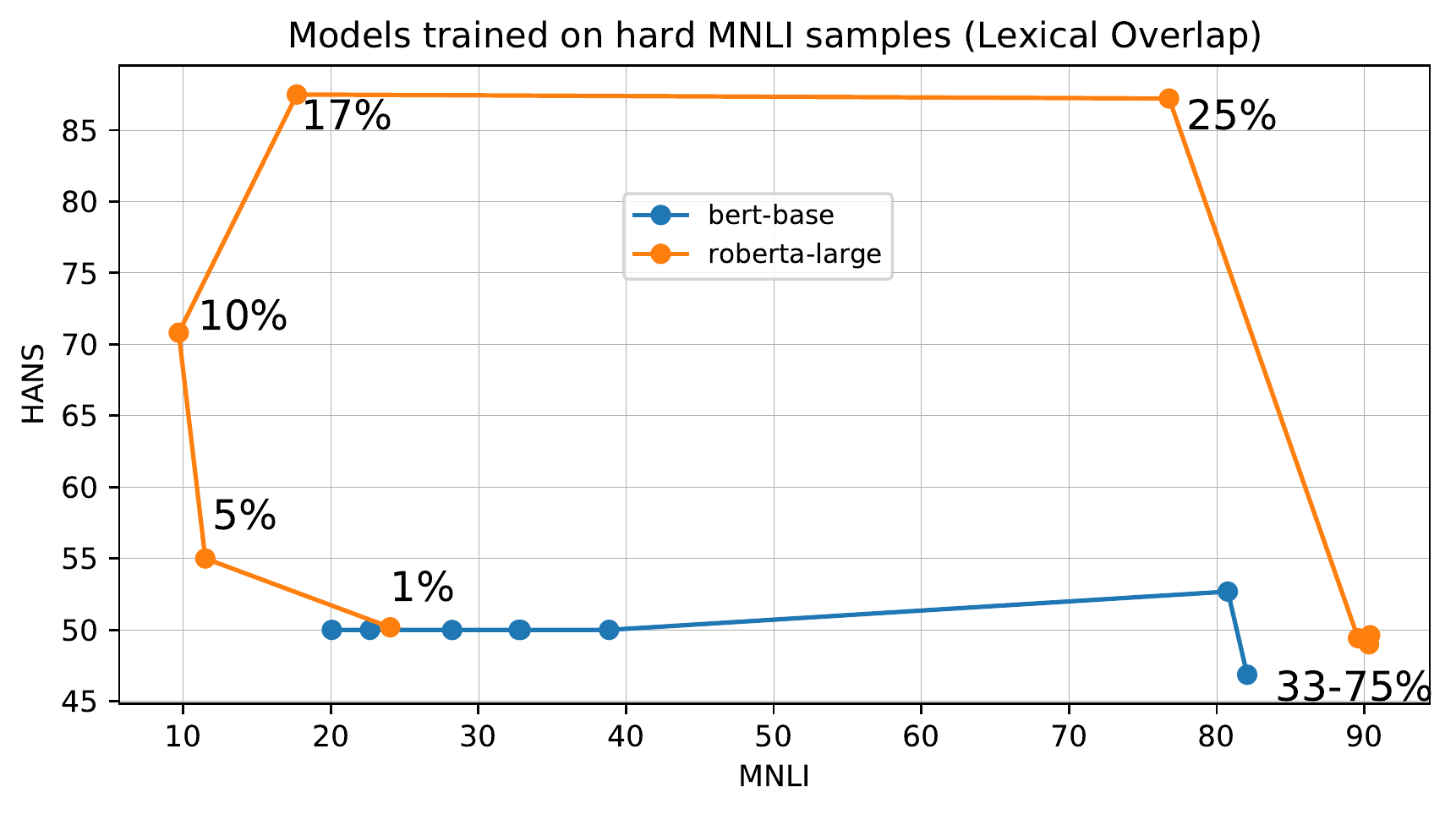}
  \caption{Generalization to HANS at varying stages of training on ``hard'' MNLI samples. Graph labels indicate \% of MNLI training data used.}
\label{fig:cartography-hard}
\end{figure}

\subsection{(Cautiously) Positive Results}
\label{sec:positive}

\Cref{fig:cartography-hard} shows that when trained on the ``hard'' samples, \textbf{for RoBERTa-large there does exist a MNLI subsample (at about 25\% training data) yielding good performance on both HANS and MNLI}. Further 13\% addition of biased MNLI data makes the model lose its performance on HANS. But we could not find such a sample for BERT-base,  although the cartography samples were model-specific. This also does not happen for either model when training on the ``ambiguous'' subsamples: RoBERTa initially ``learns'' HANS at 5\% of training data, but ``loses'' it before reaching even 60\% accuracy on MNLI (see \cref{fig:cartography-ambiguous} in the Appendix). 

\begin{figure}[!t]
\centering
    \includegraphics[width=\linewidth]{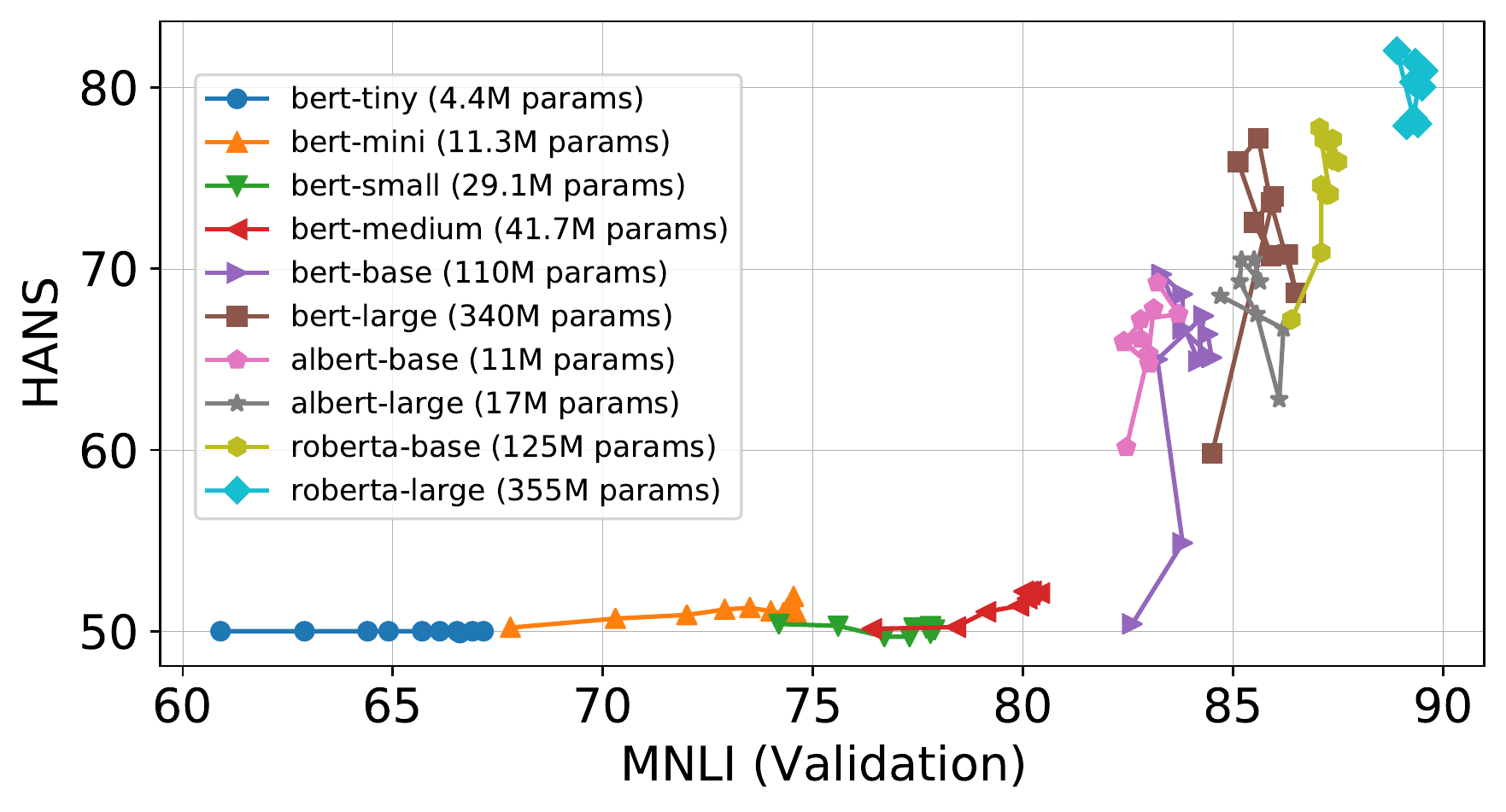}
\caption{Generalization from MNLI to HANS: model size effect (custom Pytorch Lightning implementation). The dots on the line for each model indicate performance after training epochs 1-4.}
\label{fig:bigger_models}
\end{figure}

The most encouraging results come from the increased model size with our custom trainer, as shown in \cref{fig:bigger_models}. \textbf{For BERT, }%
\textbf{RoBERTa and ALBERT, the ``large" versions generalize consistently better than the ``base" versions}. Concurrent work \cite{anonymous2021} focusing specifically on the effect of model size on the learning of lexical overlap heuristic came to a similar conclusion.

However, ``larger is better" is not the whole story. The improvement occurs only past a certain threshold: going from BERT-tiny to BERT-medium does not help generalization. At the same time, both ALBERTs have fewer parameters than BERT-small, but they do generalize, which suggests that their parameter sharing is truly effective. %
Also RoBERTa-base learns to generalize more consistently than BERT-large, which may be either due to some inherent superiority of RoBERTa, or because this instance of RoBERTa happens to be better than this instance of BERT. %
One point that is clear is that \textbf{better generalization also requires longer fine-tuning}, which interestingly barely affects the core MNLI performance on the larger models, but makes a lot of difference for the smaller BERTs. 

\section{Related work}
\label{sec:related}

Several studies have reported successful generalization from MNLI to HANS. Among data-based strategies, it has been achieved via augmenting MNLI data with predicate-argument structures \cite{MoosavideBoerEtAl_2020_Improving_Robustness_by_Augmenting_Training_Sentences_with_Predicate-Argument_Structures} and syntactic transformations \cite{min-etal-2020-syntactic}. Although there are many reports of syntactic knowledge in pre-trained BERT \cite{RogersKovalevaEtAl_2020_Primer_in_BERTology_What_We_Know_About_How_BERT_Works},  \citet{min-etal-2020-syntactic} suggest that pre-training does not yield a strong inductive bias to \textit{use} syntax in downstream tasks, and augmentation ``nudges'' the model towards that. 

Theoretically, subsampling that reduces the saliency of spurious patterns should have a similar effect, but our cartography-based subsampling did not work consistently, possibly because MNLI has little counter-evidence to spurious patterns, and the right subsample is hard to find reliably. We have additional negative results for subsampling with K-means clustering %
(see \autoref{sec:clustering} for details). %

The idea of using shallow models to identify biases \textit{before} training and ``teach'' the model to treat them differently has been successfully explored by \citet{utama-etal-2020-towards}, \citet{ClarkYatskarEtAl_2020_Learning_to_Model_and_Ignore_Dataset_Bias_with_Mixed_Capacity_Ensembles}, and \citet{SanhWolfEtAl_2020_Learning_from_others_mistakes_Avoiding_dataset_biases_without_modeling_them}. Our negative results with HEX debiasing \textit{after} training complements these reports.%

Our results corroborate that generalization is improved by larger models \cite{anonymous2021} and longer fine-tuning \cite{TuLalwaniEtAl_2020_Empirical_Study_on_Robustness_to_Spurious_Correlations_using_Pre-trained_Language_Models}. The latter work shows that this happens thanks to the few HANS-like samples in MNLI: they take longer to learn, and without them longer fine-tuning does not help. %

A general challenge for DL-based NLP research is variability due to extraneous factors. Generalization from MNLI to HANS may be much improved simply with a lucky fine-tuning initialization \cite{McCoyMinEtAl_2019_BERTs_of_feather_do_not_generalize_together_Large_variability_in_generalization_across_models_with_similar_test_set_performance}. For QA 
\citet{Crane_2018_Questionable_Answers_in_Question_Answering_Research_Reproducibility_and_Variability_of_Published_Results} show that there are many other external factors (down to linear algebra library version) that also play a role, and for Transformers overall implementations make a big difference \cite{NarangChungEtAl_2021_Do_Transformer_Modifications_Transfer_Across_Implementations_and_Applications}. Our work provides an illustration of that phenomenon in NLI. The reported HANS performance of vanilla fine-tuned BERT-base varies in the published studies from 50.0\% to 62.5\%. Our Pytorch-Lightning implementation at 4 epochs achieves 69\% (avg. of 16 runs), not due to any architectural differences. Overall it also has higher variability between runs, possibly due to batch size differences. %

\section{Analysis: Bias Under Low Confidence}
\label{sec:corruption}

Our overall ratio of positive to negative results illustrates the difficulty of the spurious patterns problem. Once the model learns that some pattern is a strong signal for a label, it will over-rely on it. But how much heuristic-matching evidence does it need?

In this experiment we fine-tune the base versions of BERT and RoBERTa for 4 epochs. We use the dataset cartography to identify the ``hard'' training samples for both models. As shown in \autoref{fig:data_map_roberta}, the classifier has overall low confidence for the ``hard'' samples. We corrupt these ``hard '' samples by inserting extra characters randomly in 30\% content words in the sequence. For example: %

{\centering
\smaller
\textit{Premise:} do it now, $\overbrace{\hbox{think'}}^{\hbox{ythink}}$ $\overbrace{\hbox{bout}}^{\hbox{ubout}}$ it later 

\textit{Hypothesis:} $\overbrace{\hbox{think}}^{\hbox{zthink}}$ about it now, do it $\overbrace{\hbox{later}}^{\hbox{late ( r}}$
}

The corrupted sequences remain relatively readable for humans, but this reduces the signal from direct lexical matches seen by the model (even with BERT tokenization). Note that the model has already seen these samples 4 times before corruption. We repeated this experiment with substituting, deleting and swapping characters.%

Since the classifier confidence for the ``hard'' examples is low, and the perturbations are random, they could be expected to just flip the predictions in random directions, equally across MNLI labels. Instead, with all corruption strategies and for both models we see the pattern shown in \autoref{tab:corruption-res}: the accuracy drops significantly for contradiction (by 10-20 points), and improves significantly for entailment (by 10-30 points). For the neutral class the change is not as large (mostly gaining 5-13 points).

These results suggest that \textbf{in low-confidence context even \textit{decreased} lexical overlap still nudges the model towards entailment} rather than contradiction. This runs contrary to the overall strategy shown by HANS, and it is not due to the majority class bias (as MNLI train set is balanced between entailment and contradiction). One possible explanation is that if it is non-entailment that the generalizing models slowly learn from the little supporting evidence in MNLI \cite{TuLalwaniEtAl_2020_Empirical_Study_on_Robustness_to_Spurious_Correlations_using_Pre-trained_Language_Models}, then corruption would make that already-difficult job even harder for the model, decreasing the accuracy on non-entailment. On the other hand, even corrupted MNLI examples still have some lexical overlap, and so the model, unable to recognize any subtler patterns, might default to that.%

This finding has implications for high-cost-of-error applications where false positives are preferable to false negatives. If the data has spurious patterns, the model may score well on a generalization benchmark, but be still biased towards a certain label when its confidence is low. Consider e.g. most of COVID detection models are ``at high risk of bias'' due to noisy data \cite{WynantsCalsterEtAl_2020_Prediction_models_for_diagnosis_and_prognosis_of_covid-19_systematic_review_and_critical_appraisal}.%

\begin{table}[!t]
\footnotesize
	\centering
	\begin{tabular}{p{1.5cm}p{1.5cm}p{1.2cm}p{1.2cm}}
		\toprule
	Corruption & Labels & BERT & RoBERTa \\ \midrule
	Character & Entailment & $+18.2$  & $+11.9$ \\
	insert & Neural & $+13.78$ & $+0.8$ \\
		& Contradiction & $-28.89$ & $-8.4$ \\
		\midrule
	Character & Entailment & $+35.5$  & $+20.4$ \\
	substitute & Neural & $+1.6$ & $+5.9$ \\
		& Contradiction & $-23.9$ & $-17.6$ \\
		\midrule
	Character & Entailment & $+23.8$  & $+18.1$ \\
	swap & Neural & $-1.6$ & $+3.3$ \\
		& Contradiction & $-15.5$ & $-13.9$ \\
		\midrule
	Character & Entailment & $+31.73$  & $+18.4$ \\
	delete & Neural & $-11.2$ & $+5.8$ \\
		& Contradiction & $-10.3$ & $-16.3$ \\
		\bottomrule
	\end{tabular}
	\caption{``Hard'' samples: changes in prediction accuracy for MNLI classes by BERT-base and RoBERTa-base after random character corruption.}
	\label{tab:corruption-res}
\end{table}

\section{Conclusion}

Most supervised datasets are biased in one way or the other, and task-independent techniques to improve NLP model generalization are crucial for further advances. We experimented with 5 strategies to improve generalization of BERT-class models for NLI task: explicit debiasing, bottlenecking the model, adapters, data subsampling, and increasing model size. We find the latter strategy the most promising, but we also report all the negative results, which contribute to the overall knowledge about generalization in BERT-based models.

\section{Acknowledgments}
This work was partially supported by JST CREST Grant Number JPMJCR19F5, Japan. We thank the anonymous reviewers for their insightful comments, T.~McCoy and T.~Linzen for the help with HANS data, and E.~Vatai for the help with langmo implementation.

\bibliography{acl2020}
\bibliographystyle{acl_natbib}

\clearpage
\appendix
\onecolumn
\section{Additional details and samples from the used datasets}
\label{app:data}
\begin{multicols}{2}  
We use MNLI to perform training of LMs and evaluate their generalization capabilities on HANS. See \autoref{tab:mnli_examples} for some sample sentences from MNLI and HANS. MNLI has three classes (``entailment", ``contradiction" and ``neutral"), while HANS only has ``entailment" and ``non-entailment". HANS targets the three heuristics (``lexical overlap", ``subsequence" and ``constituent") which are usually adopted by pre-trained LMs such as BERT.
\columnbreak

MNLI contains 393K and 20K examples in the train and dev sets respectively (the test set is not publicly available). HANS contains 30K examples split across 10K across each heuristic, which were used entirely for testing.
\end{multicols}

\vspace{0.1cm}
\begin{table}[h!]
\centering
\footnotesize
\begin{tabular}{lll} \toprule
  Label & Premise & Hypothesis \\ \midrule
  Entailment & A member of my team will - & One of our number will carry out -\\
  & execute your orders with immense precision. & your instructions minutely.\\
  
   & This information belongs to them. & How do you know? -\\
   & & All this is their information again.\\

  \midrule
  Neutral  & Product and geography are - & Conceptually cream skimming has two -\\
  & what make cream skimming work & basic dimensions - product and geography.\\
  & The speaker doesn't know who it is. & Who could there be ? \\

  \midrule
  Contradiction & I ignored Ben. & Hello, Ben. \\
  & He only muttered something about - & He distinctly said -\\
  & splitting the sky. & you were to repair the sky. \\
  \bottomrule
\end{tabular}
 
 \vspace{0.5cm}
(a) Examples of sentences from MNLI \cite{N18-1101}
 
\vspace{0.5cm}
\begin{tabular}{lllc} \toprule
  Heuristic & Premise & Hypothesis & Label \\ \midrule
  Lexical & The banker near the judge saw the actor. & The banker saw the actor. &  E  \\
  overlap & The judge by the actor stopped the banker. & The banker stopped the actor. & N \\
  \midrule
  Subsequence  & The artist and the student called the judge. & The student called the judge. & E \\
  & The senator near the lawyer danced. & The lawyer danced. & N \\
  \midrule
  Constituent & Before the actor slept, the senator ran. & The actor slept. & E  \\
  & If the actor slept, the judge saw the artist. & The actor slept. & N \\
  \bottomrule
\end{tabular} 
 
 \vspace{0.5cm}
 
(b) Examples of sentences from HANS \cite{mccoy-etal-2019-right}. \\ The \textit{label} column shows the correct label for the sentence pair; \textit{E} stands for \textit{entailment} and \textit{N} stands for \textit{non-entailment}. \\A model relying on the heuristics would label all examples as \textit{entailment} (incorrectly for those marked as N)
 
 \caption{The NLI data used in this study} %
 \label{tab:mnli_examples}
\end{table}

\twocolumn

\section{Cartography}
\label{sec:cartography-explained}
Cartography is a recent method for characterizing the difficulty of training samples. It describes data points as ``easy'', ``ambiguous'' and ``hard'' through analyzing predictions of the model during training on those samples (training dynamics). 

\textbf{Sampling.} In our experiments, we sample the datasets in the ranked order. For example, the easiest example is at the top, so it is first in the sampled batch of dataset. Our implementation of obtaining ranked ordering is based on the original implementation by the authors of the method\footnote{\url{https://github.com/allenai/cartography}}.

\begin{figure}[h!]
\centering
    \includegraphics[height=0.5\linewidth, scale=0.1]{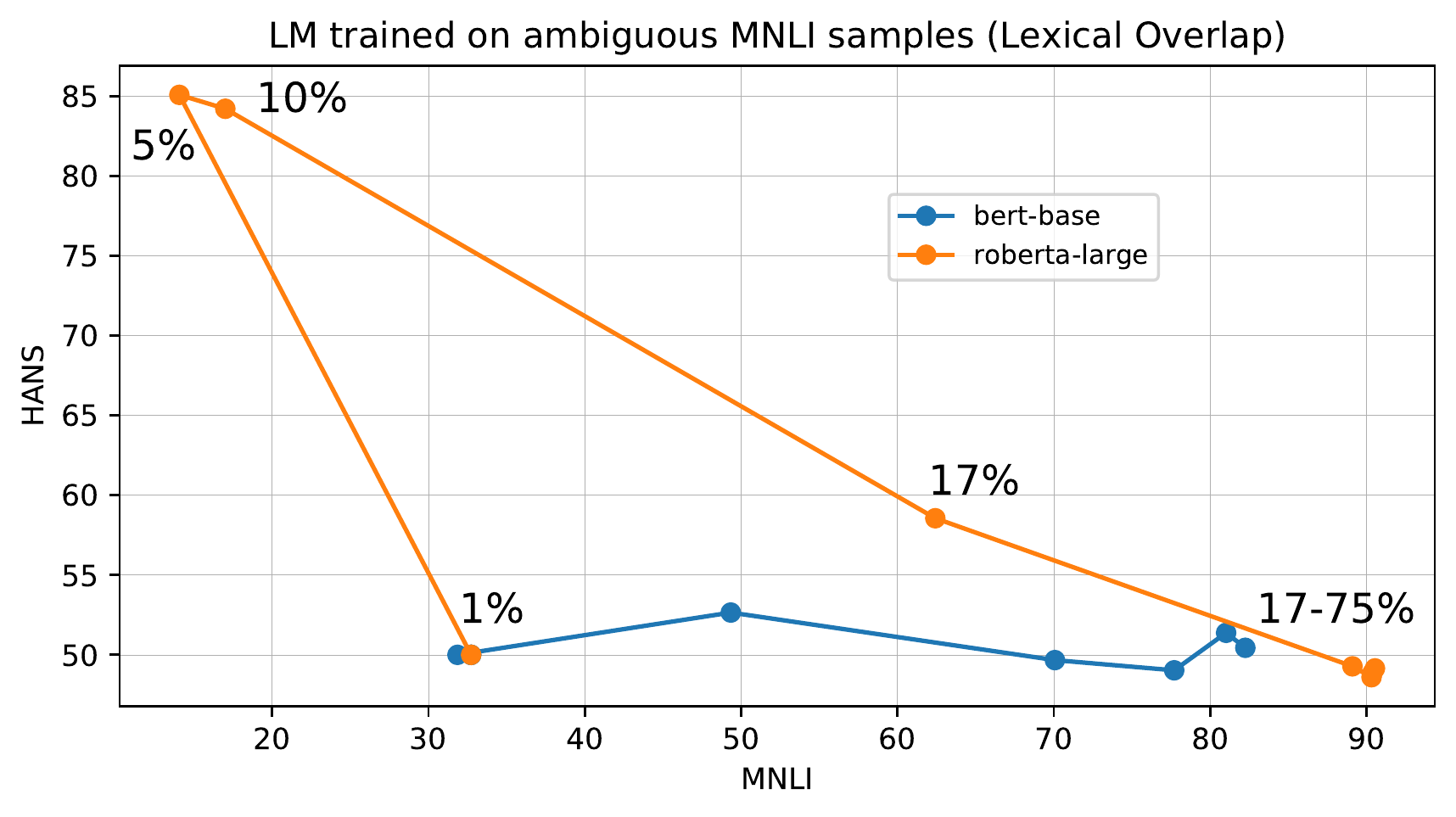}
  \caption{Evaluating generalization at varying stages of training on ``ambiguous'' samples. Percentages on marker represents percentage of MNLI train data used as training progresses.}
\label{fig:cartography-ambiguous}
\end{figure}

\textbf{Role of ``easy'' examples.} During vanilla fine-tuning on randomly subsampled data, the model encounters all three kinds of examples some of which may provide meaningful signals to learn and some may aid in out-of-domain generalization. In our experiments, we find that training solely on ``ambiguous'' and ``hard'' examples do not aid the network in improving the performance. This finding is consistent with the observation from ~\cite{swayamdipta-etal-2020-dataset} wherein they showed that ``easy'' examples aid in optimization of the network during initial stages and are crucial for training. Therefore, in our experiments the models were trained first on 25\% ``easy'' examples, and then on subsets of ``hard'' examples containing the top 1\%, 5\%, 10\%, 17\%, 25\%, 33\%, 50\% and 75\% of the ``hard'' data (for consistency with the experiment by \citet{swayamdipta-etal-2020-dataset}, see sec. 4). The results of this experiment are shown in \cref{fig:cartography-hard}. The same experiment was repeated with the ``ambiguous'' samples, shown in \cref{fig:cartography-ambiguous}. 

\section{Stabilizing HEX}
\label{appendix:hex_appendix}

\begin{figure}[!t]
\centering
    \includegraphics[height=0.6\linewidth, scale=0.7]{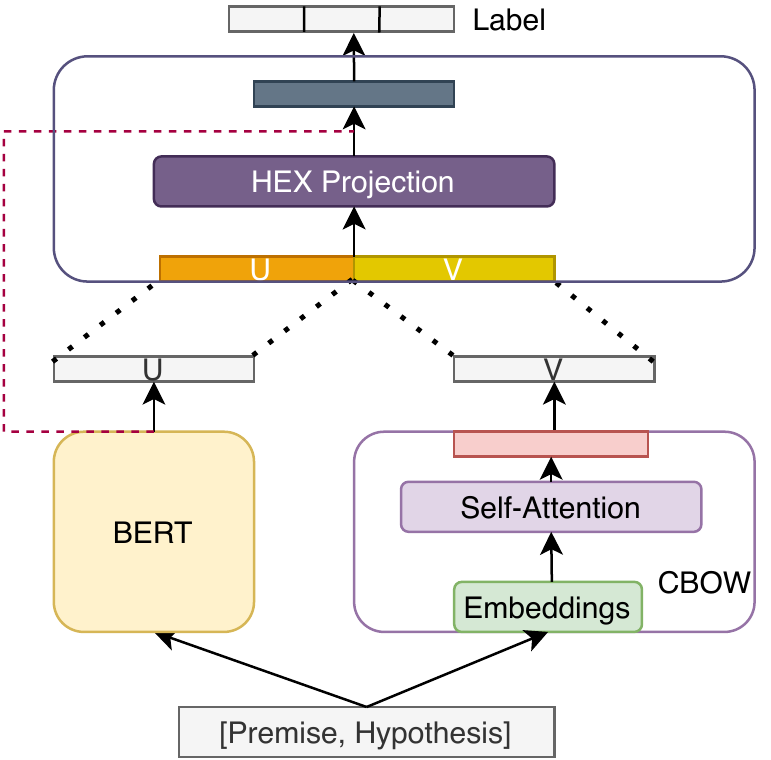}
  \caption{Orthogonal debiasing with HEX projection}
\label{fig:hex}
\end{figure}

Here we provide more details about how HEX is being used. The self-attention output from BERT is pooled and passed through two MLP layers to get an individual representation of each input sequence, as shown in \cref{fig:hex}. We feed the pooled representation of BERT and the intermediate representation of CBOW into two MLPs to obtain vectors $U$ and $V$. We use $f$ to represent classification layer parameterized by $\xi$. The output vectors $F_{A} = f([U, V]; \xi)$ and $F_{G} = f([0, V]; \xi)$ are concatenated along the non-batch dimension. 

\begin{equation}\label{eq:output_hex}
\begin{split}
F_{A} = f([U, V]; \xi), \\
F_{G} = f([0, V]; \xi) 
\end{split}
\end{equation}

where $F_{A}$, $F_{G}$ denotes both concatenated representations and zero matrix prepended with network B's representation %
$[,]$ denotes concatenation operation along the non-batch dimension. 

Following \citet{NIPS2017_7181}, we project $F_{A}$:

\begin{equation}\label{eq:hex_equation}
F_{L} = (I-F_{G}(F_{G}^{T}F_{G}+\lambda I)^{-1}F_{G}^{T})F_{A}
\end{equation}

\autoref{tab:hex_lambdas} shows hyperparameter search for $\lambda$. During inference, we use logits obtained through BERT only.%

\begin{table}[ht]
\footnotesize
\centering
    \begin{tabular}{ccc}
        \toprule
        \textbf{$\lambda$} & \textbf{MNLI} & \textbf{HANS} (Average) \\&& L\qquad S\qquad C \qquad \\
        \midrule
        1e-4 & 55.2 & 49.50 / 52.62 / 52.52\\
        2e-4 & 56.54 & 49.81 / 50.90 / 50.83 \\
        3e-4 & 57.02 & 50.03 / 50.11 / 49.93 \\
        4e-4 & 57.72 & 49.49 / 52.39 / 51.42 \\
        5e-4 & 57.09 & 49.93 / 50.16 / 50.03 \\
        6e-4 & 55.26 & 49.66 / 51.59 / 52.68 \\
        7e-4 & 57.25 & 49.60 / 51.09 / 49.63 \\
        8e-4 & 57.78 & 49.60 / 51.20 / 51.37 \\
        9e-4 & 48.58 & 50.00 / 50.00 / 49.98 \\
        1e-5 & 53.45 & 49.80 / 50.69 / 50.68 \\
        2e-5 & 53.77 & 50.00 / 50.00 / 50.00 \\
        3e-5 & 56.46 & 49.19 / 52.84 / 54.41 \\
        4e-5 & 47.30 & 49.61 / 50.90 / 49.28 \\
        5e-5 & 50.80 & 49.69 / 51.07 / 50.41 \\
        \bottomrule
    \end{tabular}
    \caption{Performance on MNLI and HANS with HEX (BERT-base) with different values of $\lambda$. 
    L, S, C denote lexical overlap, subsequence and constituent heuristic} 
    \label{tab:hex_lambdas}
\end{table}

\begin{figure*}[tp]
    \begin{subfigure}[b]{0.31\textwidth}
        \includegraphics[height=3.5cm]{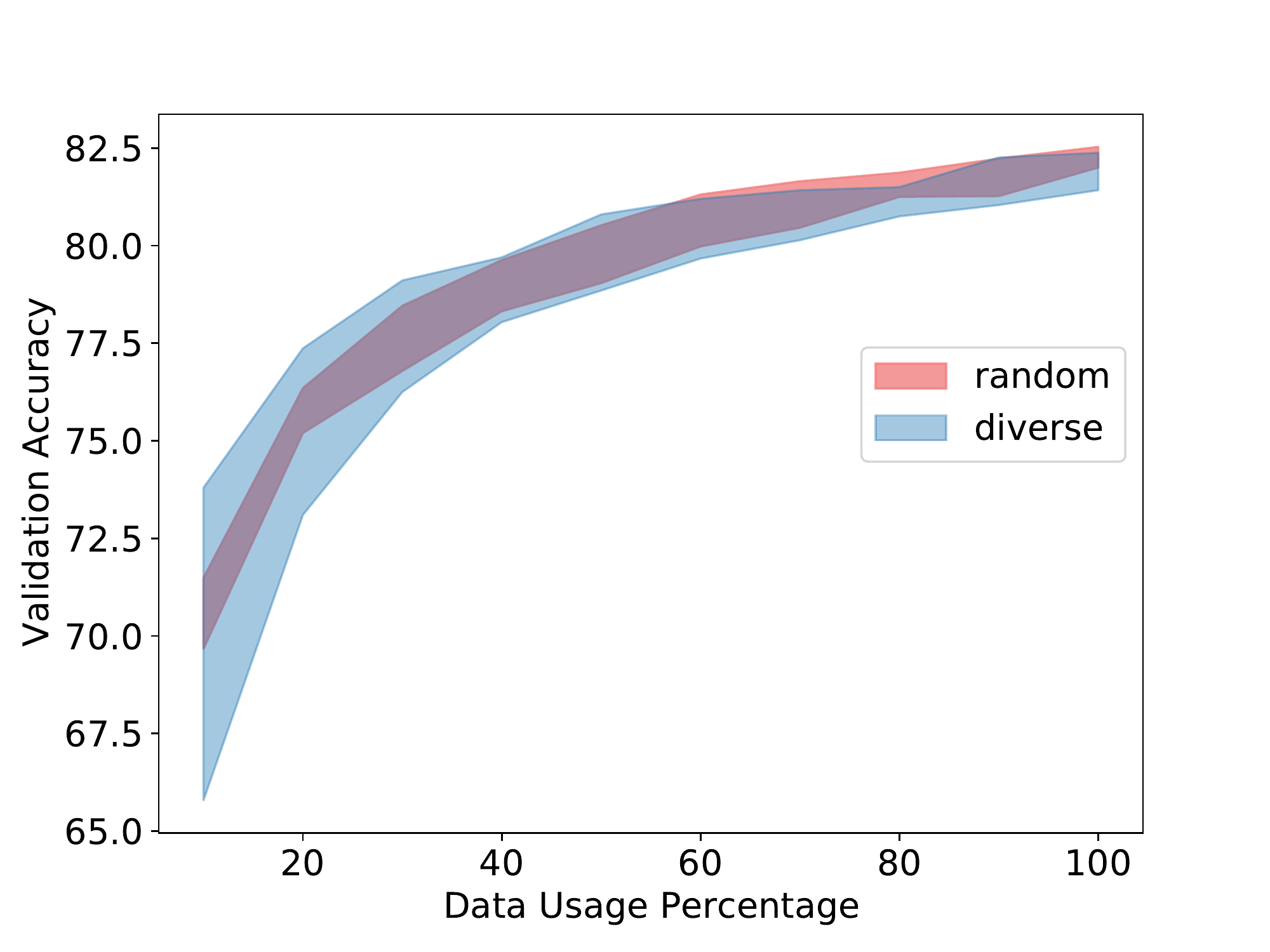}
        \caption{Diverse vs Random sampling: MNLI}
        \label{fig:sampling_mnli}
    \end{subfigure}
    \hfill
    \begin{subfigure}[b]{0.31\textwidth}
        \includegraphics[height=3.5cm]{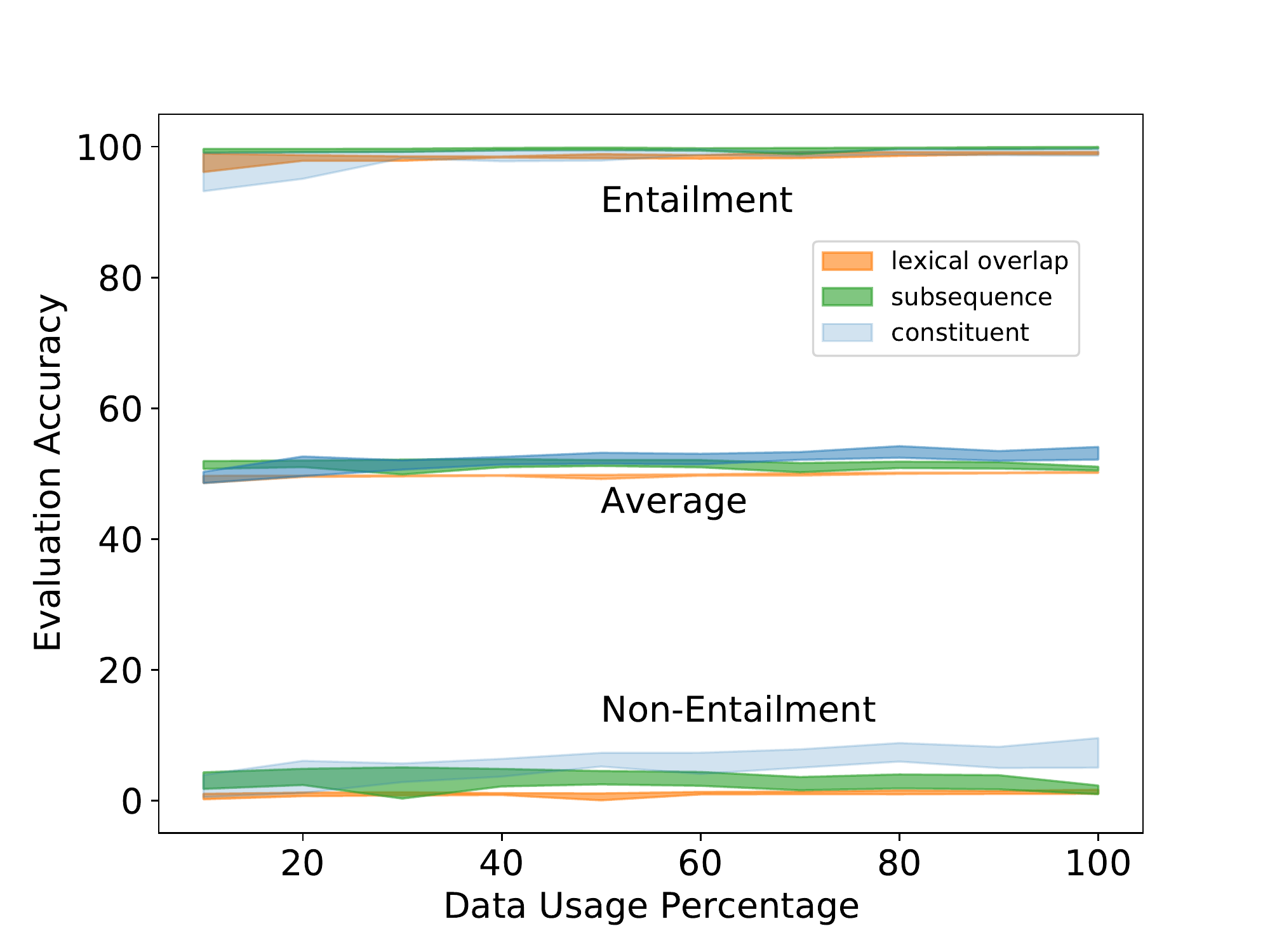}
        \caption{Random sampling: HANS}
        \label{fig:sampling_hans_random}
    \end{subfigure}
    \hfill
    \begin{subfigure}[b]{0.31\textwidth}
        \includegraphics[height=3.5cm]{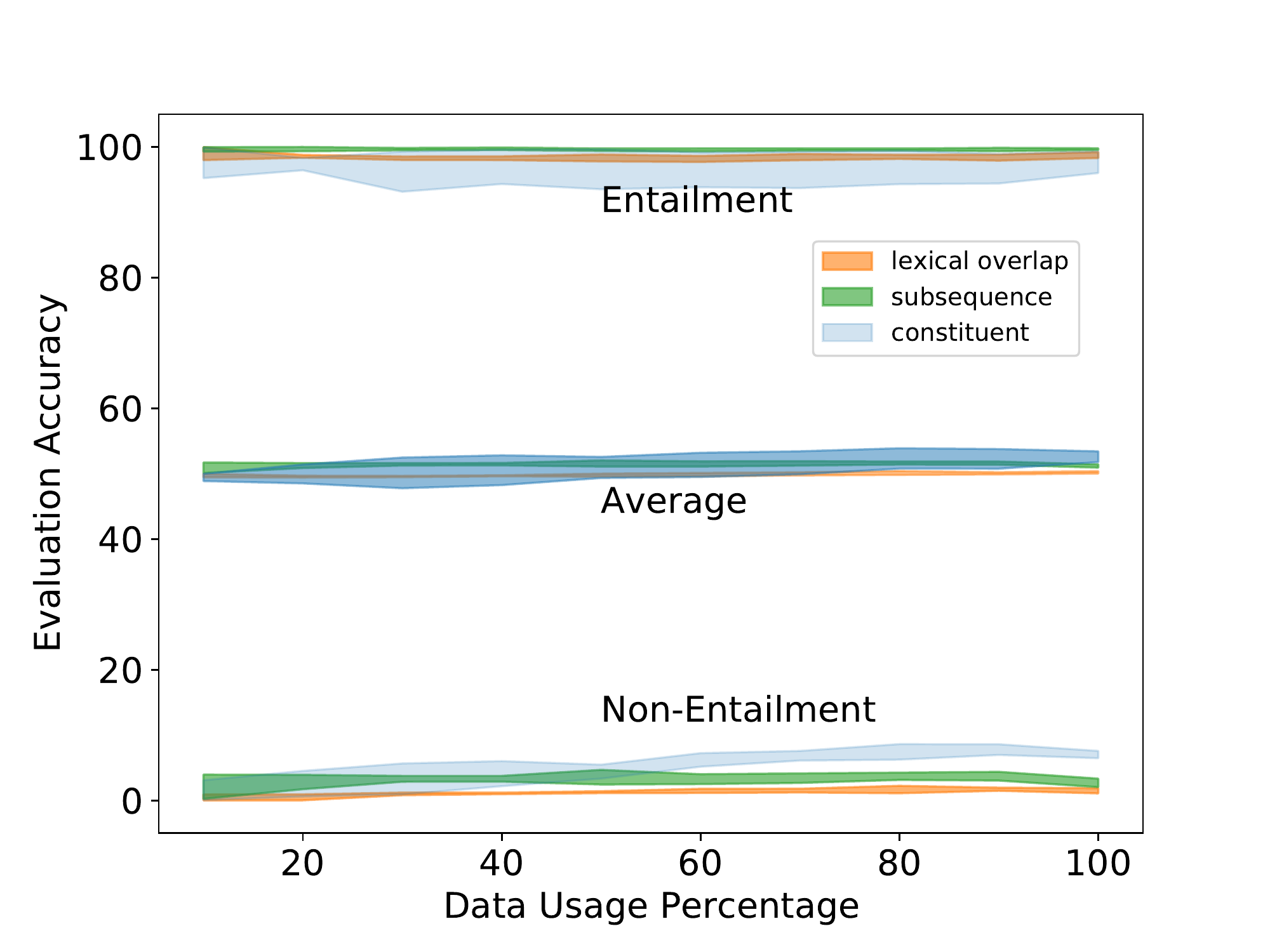}
        \caption{Diverse sampling: HANS}
         \label{fig:sampling_hans_diverse}
    \end{subfigure}
\caption{Fine-tuning BERT-base on varied amounts of MNLI data: in-domain and generalization performance}
\label{fig:progressive_subsampling}
\end{figure*}

We follow a slight variation of \autoref{eq:hex_equation} to smoothen the process of optimization. The addition of $\lambda$ hyper-parameter has been done to ensure that inverse is carried out on a non-singular matrix. The value of $\lambda$ plays a significant role in determining how these representations are being learned. In our experiments, $1e-4$ worked well and was used for initializing it after which it was set as a model's parameter. We observe that pseudo-inverse of $F_{G}^{T}F_{G}$ is unstable and can make optimization process hard, so we make $U$ and $V$ square matrices to obtain inverse instead. Additionally, during inference time, we directly feed outputs from the main network to the MLP layer to obtain logit vectors instead of using $F_{L}$. It has been reported \cite{wang2018learning} that this doesn't have any profound impact on the logit vector and makes inference faster. We also applied L1 and L2 normalization on $U$ and $V$ to account for differences in scale but did not see any noticeable improvement. We found that values of $\lambda$ greater than $0.0001$ do not aid the network in learning.

\section{Subsampling the Training Data with K-means clustering}
\label{sec:clustering}

\textbf{Motivation.} Fundamentally, the problem is the mismatch between MNLI and HANS distributions. %
For a biased dataset, one solution could be to find such a subsample that would enable the model to perform well on both distributions.%

\textbf{Experiments.} We encode MNLI examples as BERT [CLS] embeddings and cluster them in 512 clusters using K-means. We then fine-tune BERT-base on varying amounts of MNLI data, progressively increasing the amount of training examples by 10\%. The data in the sub-sample is selected (a) randomly (as a control), (b) so as to maximize the diversity of examples within the sample \cite{DBLP:conf/icml/KatharopoulosF18}.  At the smallest subsample size we sample the data from all clusters. As data size increases, smaller clusters are exhausted while the larger ones remain, so the smallest subsamples are the most diverse, and the diversity decreases as the sample size increases. The experiment is repeated with 5 random seeds.

\textbf{Results.} 
\autoref{fig:progressive_subsampling} shows that on MNLI, diverse sampling yields much more variation with small amounts of data than random sampling, but as the subsample approaches the full dataset the performance also becomes the same. Neither subsampling strategy improves generalization: the model still predicts ``entailment" for most HANS examples. Thus overall the result for this strategy is negative.%

\cleardoublepage

\end{document}